\documentclass{article}

\PassOptionsToPackage{numbers, compress}{natbib}

\usepackage[preprint]{neurips_2024}

\usepackage[utf8]{inputenc} 
\usepackage[T1]{fontenc}    
\usepackage{hyperref}       
\usepackage{url}            
\usepackage{booktabs}       
\usepackage{amsfonts}       
\usepackage{nicefrac}       
\usepackage{microtype}      
\usepackage{xcolor}         
\usepackage{amsmath}
\usepackage{graphicx}
\usepackage{multirow}
\usepackage{float,caption}
\usepackage{wrapfig}

\title{FFAM: Feature Factorization Activation Map for Explanation of 3D Detectors}

\author{
Shuai~Liu,~Boyang~Li,~Zhiyu~Fang,~Mingyue~Cui,~Kai~Huang\thanks{Corresponding author.} \\
School of Computer Science and Engineering, Sun Yat-sen University\\
\{liush376@mail2, liby83@mail, fangzhy9@mail2, cuimy@mail2, huangk36@mail\}.sysu.edu.cn
}

\begin{document}

\maketitle

\begin{abstract}
LiDAR-based 3D object detection has made impressive progress recently, yet most existing models are black-box, lacking interpretability. Previous explanation approaches primarily focus on analyzing image-based models and are not readily applicable to LiDAR-based 3D detectors. In this paper, we propose a \textit{feature factorization activation map} (FFAM) to generate high-quality visual explanations for 3D detectors. FFAM employs non-negative matrix factorization to generate concept activation maps and subsequently aggregates these maps to obtain a global visual explanation. To achieve object-specific visual explanations, we refine the global visual explanation using the feature gradient of a target object. Additionally, we introduce a voxel upsampling strategy to align the scale between the activation map and input point cloud. We qualitatively and quantitatively analyze FFAM with multiple detectors on several datasets. Experimental results validate the high-quality visual explanations produced by FFAM. The Code will be available at \url{https://github.com/Say2L/FFAM.git}.

\end{abstract}

\section{Introduction}

In recent years, there has been rapid development in LiDAR-based 3D object detection \cite{second, dsvt, club, hednet, dcdet}, making it widely utilized in autonomous driving, industrial automation, and robot navigation. However, existing detection methods predominantly rely on deep neural networks with highly nonlinear and complex structures. Essentially, these models can be considered as "black box" systems. Such opaque modeling techniques hinder users from fully trusting the detection models, particularly in sensitive and high-risk domains. Consequently, understanding the decision-making process of these inherently opaque models is urgently needed.

Visual explanation methods \cite{gradcam, gradcam++, scorecam, rise, drise} have gained widespread adoption for analyzing models based on deep neural networks. These methods generate saliency maps that highlight the crucial elements influencing the model's decision within the input map. Perturbation-based \cite{rise, drise}, class activation map (CAM)-based \cite{cam, gradcam, gradcam++}, and gradient-based \cite{ig, smoothgrad, fullgradient} methods are the three main categories of visual explanation methods. However, these methods primarily focus on image-based models and are not directly applicable to point cloud-based models. The pioneering work in analyzing 3D detectors is OccAM \cite{occam}, which extends D-RISE \cite{drise} to perturb point clouds. As a perturbation-based approach, OccAM first randomly samples numerous sub-point clouds and measures the change in model predictions. However, the large number of inference calculations makes OccAM computationally intensive, and the sampling number easily impacts the quality of generated saliency maps.

Interpreting 3D detectors presents three key challenges. First, point clouds are inherently three-dimensional (3D). It is essential to generate corresponding 3D saliency maps for accurate interpretation. However, existing methods, such as popular CAM-based techniques, primarily utilize activation maps from the network's last layer to generate 2D saliency maps. Second, the explanation method for 3D detectors should provide detailed explanations for individual objects of interest. Yet, most existing methods yield class-specific saliency maps, which means they cannot focus on explaining a specific detection object. Lastly, point clouds are sparsely distributed in 3D space, rendering linear interpolation employed by many image-based explanation methods ineffective. 

\begin{wrapfigure}{r}{0.6\textwidth} 
  \centering
  \includegraphics[width=0.6\textwidth]{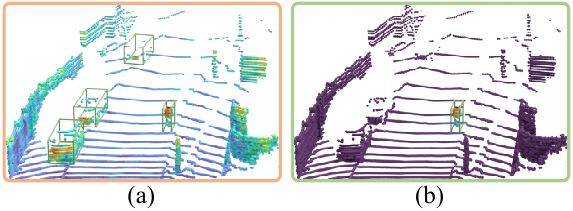} 
  \caption{Visualization of FFAM outputs. (a) global concept activation map and (b) object-specific activation map. }
  \label{ffam_demo}
\end{wrapfigure} 


To address the aforementioned challenges, this paper introduces a feature factorization activation map (FFAM) to obtain visual explanations for 3D detectors. Specifically, to solve the first challenge, FFAM leverages the 3D feature maps within the 3D backbone \cite{second} of detectors, rather than relying on the bird's eye view (BEV) feature maps from the last layer. Drawing inspiration from DFF \cite{dff}, we employ non-negative matrix factorization (NMF) to uncover latent semantic concepts within these 3D feature maps. Typically, point features with effective detection clues in 3D detectors contain richer semantic concepts. Thus, we aggregate the concept activation maps to generate a global concept activation map that highlights important points, as shown in Figure~\ref{ffam_demo}(a). To address the second challenge of obtaining object-specific saliency maps, we utilize the gradients of the 3D feature map, generated by an object-specific loss, to refine the global concept activation map. This process is illustrated in Figure~\ref{ffam_demo}(b), showcasing the desired effect. To tackle the final challenge, we introduce a voxel upsampling strategy to sample values from sparse neighbors, ensuring accurate saliency map generation. 

We compare our FFAM with the current state-of-the-art method OccAM \cite{occam}, as well as other image-based explanation methods including Grad-CAM \cite{gradcam} and ODAM \cite{odam}. We conduct experiments on the KITTI \cite{kitti} and Waymo Open \cite{waymo} datasets, employing detectors such as SECOND \cite{second} and CenterPoint \cite{centerpoint}. The qualitative and quantitative results demonstrate that our FFAM significantly outperforms the previous methods. The contributions of this work can be summarized as follows:
\begin{itemize}
    \item We propose a feature factorization activation map (FFAM) method to obtain high-quality visual explanations for 3D detectors. 
    
    \item We first introduce NMF in explaining point cloud detectors. By aggregating different concept activation maps, we obtain a global concept activation map that highlights points with significant detection clues.
    
    \item We utilize feature gradients of an object-specific loss to refine the global concept activation map, enabling the generation of object-specific saliency maps.
    
    \item A voxel upsampling strategy is proposed to upsample sparse voxels, thus aligning the scale between the activation map and input point cloud. 
\end{itemize}

\section{Related Work}
\noindent\textbf{Explanation Methods for Image-based Models. }Existing explanation methods primarily focus on image classification models. Perturbation-based methods \cite{sun2020explaining, rise, dabkowski2017real, wagner2019interpretable} are widely used for interpreting image classification models. The core idea is to assign importance scores to perturbed feature components by disturbing the model's input and observing the output changes. CAM-based methods \cite{cam, gradcam, gradcam++, layercam} generate saliency maps by linearly combining activation maps from intermediate layers, weighted by their respective contributions. Some approaches (e.g. Score-CAM \cite{scorecam} and Ablation-CAM \cite{ablationcam}) combine perturbation- and CAM-based ideas to eliminate dependence on backpropagation gradients. Additionally, gradient-based explanation methods \cite{simonyan2013deep, ig, smoothgrad, fullgradient} use gradients to quantify input impact on network predictions. Higher gradient values indicate greater importance of the corresponding input elements. Moreover, feature factorization techniques like principal component analysis (PCA) and non-negative matrix factorization (NMF) can uncover latent patterns in deep features. DFF \cite{dff} employs NMF to localize semantic concepts within images.


Compared to explanation methods for classifiers, only a limited number of approaches investigate explanations for object detection models. The aforementioned methods generate class-specific explanations, which are not feasible for object detection models. D-RISE \cite{drise} employs a perturbation strategy to generate instance-specific explanations by defining a detection similarity metric. In \cite{wu2019towards}, a directed acyclic AND-OR Graph (AOG) is utilized to uncover latent structures in object detectors. G-CAME \cite{gcame} combines activation maps with a Gaussian kernel of gradients to generate a saliency map for a predicted bounding box. ODAM \cite{odam} employs pixel-wise gradients of a target object to weigh the activation maps, thereby producing an instance-specific saliency map.

\noindent\textbf{Explanation Methods for Point Cloud-based Models. }
In contrast to explanation methods for image-based models, the field of explanation for point cloud-based models is relatively underdeveloped. Existing methods primarily focus on point cloud classification models. For instance, \cite{zheng2019pointcloud} utilizes the loss gradient to measure the contribution of each point in the classifier. Similarly, \cite{gupta20203d} applies a gradient-based strategy to analyze the intermediate features of the network. Another approach \cite{tan2023visualizing} combines a generative model with the activation maximization method \cite{erhan2009visualizing} to obtain a global explanation for point cloud networks.


Research on the explanation of 3D detectors is still quite limited. One perturbation-based method, OccAM \cite{occam}, estimates the importance of individual points by testing the model with randomly generated subsets of the input point cloud. However, the scale of points in 3D space is considerably large, and the distribution of points acquired through LiDAR varies with distance. These aforementioned issues result in the following challenges for perturbation-based methods: (1) It is difficult to exhaustively perturb the point cloud, limiting the quality of visual explanations; (2) Generating ample random subsets of points requires multiple iterations, thereby reducing efficiency. Taking inspiration from feature factorization techniques \cite{dff} and gradient-based approaches \cite{ig, gcame, odam}, we propose an explanation method called FFAM. It aims to efficiently generate high-quality saliency maps for 3D detectors.

\noindent\textbf{LiDAR-based 3D Object Detection. }
These methods can be categorized into two main groups: one-stage and two-stage detectors. \textbf{One-stage detectors} typically employ simple network architectures to achieve high speeds. For instance, SECOND \cite{second} efficiently encodes sparse voxel features using a proposed 3D sparse convolution technique. PointPillars \cite{pointpillars} divides a point cloud into pillar voxels, eliminating the need for 3D convolution layers and achieving fast inference speed. VoxelNeXt \cite{voxelnext} introduces a fully sparse convolution network that eliminates the requirement for sparse-to-dense conversion. \textbf{Two-stage detectors} generally incorporate an additional stage to refine proposals generated by a one-stage network. PointRCNN \cite{pointrcnn} utilizes PointNet++ \cite{pointnet++} to generate proposals from raw points and then refines the bounding boxes in the second stage. PV-RCNN \cite{pv-rcnn} combines a voxel-based proposal network with a point-based refinement network. CenterPoint \cite{centerpoint} extracts point features from the surface centers of proposal bounding boxes for refinement. Voxel R-CNN \cite{voxelrcnn} utilizes voxel features from the 3D backbone to refine the proposals. Our explanation method FFAM is adaptable to both one- and two-stage detectors without being limited by the detector type. We primarily conduct experiments on widely used detectors, including the one-stage detector SECOND and the two-stage detector CenterPoint.

\section{Method}
The goal of visual explanation for a 3D detector $f$ is to produce a saliency map for each detection. Given a point cloud $P \in \mathbb{R}^{N\times4}$, the saliency map consists of $N$ values presenting the importance of each point in $P$ for a detection $d$ which consists of a bounding box, a confidence score and a category label. We denote a detection $d$ as follows:

\begin{figure*}[t]
    \centering
    \includegraphics[width=0.9\textwidth]{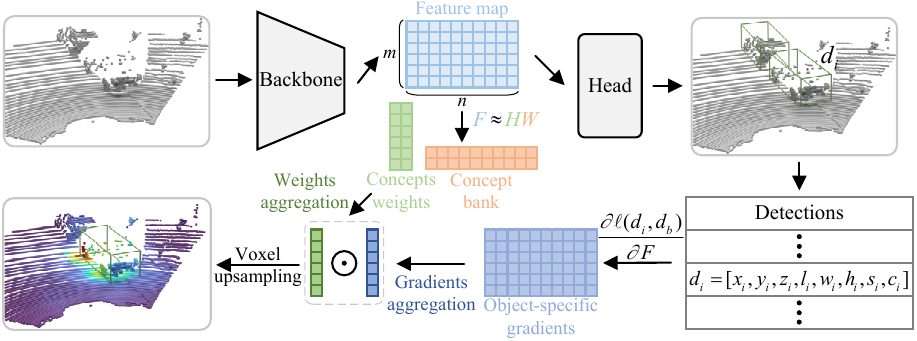}
    \caption{Overall framework of our FFAM which can generate an object-specific saliency map for a detection $d_i$. }
    \label{method_overview}
\end{figure*}

\begin{equation}
    d = \left[x, y, z, l, w, h, s, c \right]
\end{equation}

\noindent where $(x, y, z)$ denotes the center location, $(l, w, h)$ represents the object size (i.e., length, width and height), $s$ and $c$ indicate the confidence score and category label, respectively. 

We propose FFAM to produce saliency maps in point cloud format for 3D detectors. The overview of our method is illustrated in Figure~\ref{method_overview}. It can be divided into three phases as follows: (1) Feature factorization (Sec.~\ref{feature factorization}); (2) Gradient weighting (Sec.~\ref{gradient weighting}); (3) Voxel upsampling (Sec.~\ref{voxel upsampling}).

\subsection{Feature Factorization Activation Map}
\label{feature factorization}

Matrix factorization is widely used in fields such as recommendation systems, image processing, and natural language processing to extract potential features and reduce dimensionality. Non-negative matrix factorization (NMF) as a classical matrix factorization algorithm approximates a non-negative matrix by decomposing it into the product of two non-negative matrices. With this decomposition, NMF can discover potential patterns and conceptions in the raw matrix and extract the most important features. Given a non-negative matrix $A \in \mathbb{R}^{m \times n}$, NMF retrieves an approximation $\hat{A} \in \mathbb{R}^{m \times n}$ as follows:

\begin{equation}
    \begin{aligned}
        \text{NMF}(A)  = \underset{\hat{A}}{\operatorname{argmin}}\left\|A-\hat{A}\right\|_{F}^{2}, \\
        \text{s.t. } \hat{A} = HW, \forall ij, H_{i j}, W_{i j} \geq 0,
    \end{aligned}
    \label{Eq:eq6}
\end{equation}

\noindent where $H \in \mathbb{R}^{m \times r}$ and $W \in \mathbb{R}^{r \times n}$ denote two non-negative matrics. $r$ is a predefined parameter indicating the number of latent concepts in matrix A. Each row $W_j \in \mathbb{R}^n (1 \le j \le r)$ of $W$ represents a concept vector. These concept vectors are typically well-interpreted and associated with object-part features, such as wheels, car doors, car roofs, and so on, following the non-negative additivity property of $W_j$. Furthermore, each row $H_i \in \mathbb{R}^{r}$ (where $1 \le i \le m$) of matrix $H$ represents the combination weights of different concept vectors in $W$. Combining these concept vectors using the weights $H_i$, we obtain the $i$-th row feature of matrix $\hat{A}$.

In this paper, we employ non-negative matrix factorization to handle the voxel feature map within the 3D backbone of detectors. Typically, voxel features that contain crucial detection clues tend to activate more concepts (e.g., license plates, car fronts, car edges) in detectors. As a result, aggregating all weights in $H_i$ indicates the significance of the $i$-th voxel feature in the voxel feature map, as demonstrated in Figure~\ref{ffam_demo}(a). 

Specifically, given a voxel feature map $F \in \mathbb{R}^{M \times d}$ where $M$ represents the voxel number and $d$ denotes the channel number, a voxel feature $F_i \in \mathbb{R}^d$ in $F$ can be factorized as follows:

\begin{equation}
    \begin{aligned}
        F_i = \sum_{j=1}^{r}H_{ij}W_j.
    \end{aligned}
\end{equation}

Further, we obtain the global concept activation map $V$ by aggregating concept weight matrix $H$ as follows:

\begin{equation}
    \begin{aligned}
        V = \sum_{j=1}^{r}H_{\cdot j},
    \end{aligned}
\end{equation}

\noindent where $H_{\cdot j}$ denotes $j$-th column of $H$. The resulting $V$ emphasizes points with multiple activated concepts from a global perspective. And due to the downsampling operation in the detection network, the granularity of $V$ is typically coarse. Therefore, further processing is required to obtain an object-specific and fine-grained activation map, as described in Sec.~\ref{gradient weighting} and Sec.~\ref{voxel upsampling}.

\subsection{Object-Specific Gradient Weighting}
\label{gradient weighting}
In a 3D detector, the output contains a large number of detections. To obtain an object-specific activation map, we establish a loss function for a specific detection. Specifically, given a detection $d$, we create a baseline detection $d_b$ to calculate the loss $\ell$:

\begin{equation}
    \begin{aligned}
        \ell = \left \|  d - d_b\right \| _1.
    \end{aligned}
\end{equation}

For simplicity, we use the L1 loss function and set all values in $d_b$ equal to 0. Then we obtain the gradient map $G \in \mathbb{R}^{M \times d}$ of the feature map $F$:

\begin{equation}
    \begin{aligned}
        G = \frac{\partial \ell}{\partial F}.
    \end{aligned}
\end{equation}

Considering an optimization process from $d$ to $d_b$, the matrix $G$ denotes the optimal direction for reducing the loss. If we iteratively update the feature map $F$ based on the gradient map $G$, the information related to the detection $d$ will be diminished. Alternatively, by utilizing $G$, we can identify the locations in the feature map $F^l$ that contain clues about $d$. Consequently, an object-specific activation map $M$ for $d$ can be obtained as follows:

\begin{equation}
    \begin{aligned}
        \omega = \sum_{k=1}^{d} \left| G_{\cdot k} \right|, \\
        M = \Phi(\omega) \odot \Phi(V),
    \end{aligned}
\end{equation}


\noindent where $G_{\cdot k} \in \mathbb{R}^M$ refers to the $k$-th column of $G$, while $\Phi$ represents the normalization operation, and $\odot$ denotes element-wise multiplication. By modifying the loss function to a specific attribute $p$ in detection $d$, we can examine the specific points on which the detector concentrates when predicting attribute $p$.

\subsection{Voxel Upsampling}
\label{voxel upsampling}

Due to downsampling operations in 3D detection networks, the scale of the activation map is typically smaller than that of the input point cloud. Consequently, upsampling the activation map $M$ becomes necessary. However, unlike 2D images, linear interpolation for upsampling 3D sparse voxels presents challenges. To address this, we draw inspiration from the voxel query technique proposed by \cite{voxelrcnn} and introduce a voxel upsampling strategy for 3D sparse voxels. Specifically, we define the voxel size as $s$, and the ranges of the point cloud for three axes as $[x_l, x_r]$, $[y_l, y_r]$, and $[z_l, z_r]$ respectively. Given a point $p=(x, y, z)$, we calculate the coordinate $(x_p, y_p, z_p)$ of voxel $v_p$ to which $p$ belongs as follows:

\begin{equation}
    \begin{aligned}
        x_p=\left \lfloor  \frac{x - x_l} {s}\right \rfloor, \text{ } y_p = \left \lfloor  \frac{y - y_l} {s}\right \rfloor, \text{ }z_p = \left \lfloor  \frac{z - z_l} {s}\right \rfloor.
    \end{aligned}
\end{equation}

Then we query neighbor voxels on activation map $M$ for $p$, using Manhattan distance to control the query range:

\begin{equation}
    \begin{aligned}
        d(v_p, v) = |x_n - x_p| + |y_n - y_p| + |z_n - z_p|,
    \end{aligned}
\end{equation}

\noindent where $(x_n, y_n, z_n)$ is the coordinate of an neighbor voxel $v$, $d(\cdot, \cdot)$ is the Manhattan distance between two voxels. We sample up to $k$ neighbor voxels within a distance threshold. Finally, the salience score $s_p$ of point $p$ is calculated as follows:

\begin{equation}
    \begin{aligned}
        s_p = \sum_{v \in \aleph} \frac{\Psi (d(v_p, v))}{\sum_{v \in \aleph} \Psi (d(v_p, v))}  M_v,
    \end{aligned}
\end{equation}

\noindent where $\aleph$ is the set of neighbor voxels, $\Psi$ denotes a Gaussian kernel with a standard normal distribution, $M_v$ represents the value of voxel $v$ on activation map $M$.

\begin{figure*}[t]
    \centering
    \includegraphics[width=1.0\textwidth]{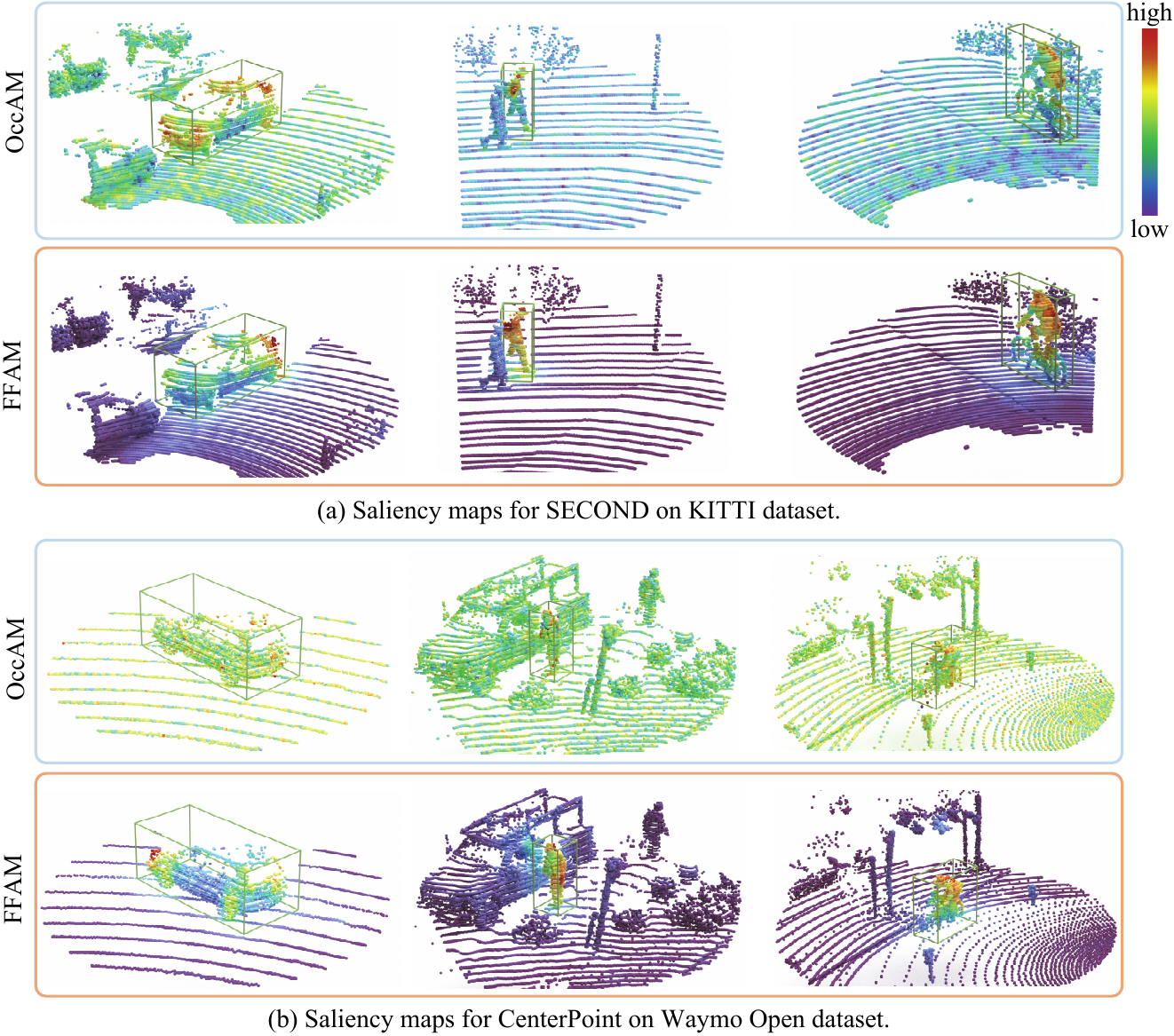}
    \caption{Saliency maps for SECOND \cite{second} and CenterPoint \cite{centerpoint}. The green bounding boxes indicate the detected objects, while warmer colors (using the turbo colormap) represent higher point contributions to these detections. The crops are provided for visualization purposes only.}
    \label{qualitative_comparison}
\end{figure*}

\section{Experiments}
In this section, we compare our FFAM with existing explanation methods, including Grad-CAM \cite{gradcam} and ODAM \cite{odam}, for image-based models, as well as with OccAM \cite{occam}, the state-of-the-art explanation method for point cloud-based models. We adopt two datasets for evaluation: KITTI \cite{kitti}, a widely used autonomous driving dataset, and Waymo Open \cite{waymo}, containing complex multi-object scenes. For KITTI, experiments are conducted on SECOND \cite{second}. For Waymo Open, we mainly evaluate on CenterPoint \cite{centerpoint}. The experiments are run using PyTorch and an RTX 3090 GPU. The hyperparameters of detectors and OccAM remain consistent with their official implementations. The parameter $r$ used in NMF is set to 64. The Manhattan distance threshold and parameter $k$ in voxel upsampling are set to 2 and 16, respectively. We use the 3D feature map from the third block of the 3D backbone as FFAM input. Hyperparameters analysis and ablation study are in App.~\ref{hyperp_analysis} and App.~\ref{ablation}, respectively.


\subsection{Qualitative Results}
To verify the interpretability of our FFAM, we visualize explanations for some objects. We also visualize the average saliency maps of different categories for specific object attributes to study the latent pattern of 3D detectors.

\textbf{Visualization of Saliency Map.} 
We compare the visual explanations generated by FFAM and OccAM \cite{occam} for cars, pedestrians, and cyclists in Figure~\ref{qualitative_comparison}(a). These detection results are obtained by SECOND \cite{second} detector trained on KITTI \cite{kitti}. OccAM exhibits significant background noise due to its random masking mechanism. In contrast, our ODAM demonstrates a strong ability to generate clear, distinct object-specific saliency maps. We observe the detector also captures relevant clues from the background and neighboring objects. Furthermore, we compare saliency maps generated by FFAM and OccAM on Waymo Open \cite{waymo} using the CenterPoint \cite{centerpoint} detector, as shown in Figure~\ref{qualitative_comparison}(b). The saliency maps produced by OccAM struggle to focus on the intended object for interpretation. They have more highly salient points distributed on the background compared to KITTI. We attribute this discrepancy to the larger number of points in Waymo Open samples, challenging the random masking mechanism to sample diverse point masks effectively. Conversely, our FFAM consistently generates high-quality saliency maps on Waymo Open.

\begin{figure}[t]
    \centering
    \includegraphics[width=1.0\textwidth]{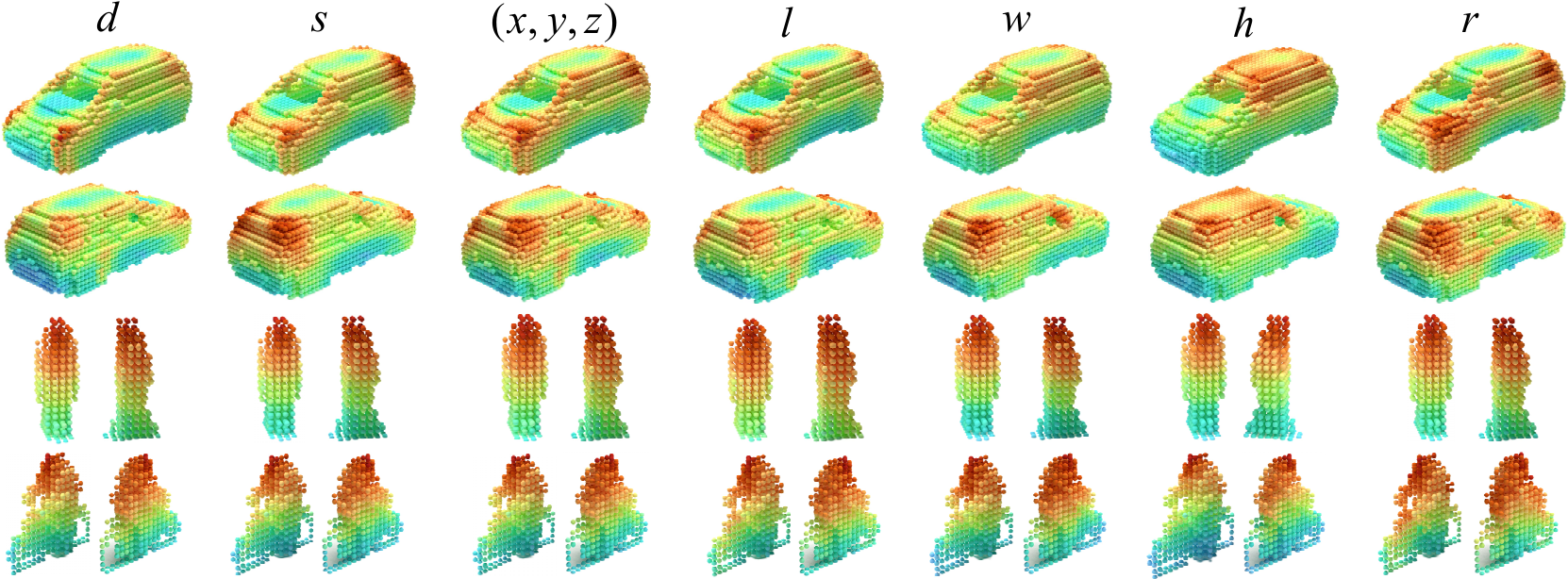}
    \caption{Average saliency maps for different object attributes. $(x,y,z)$ denotes the center of predicted object. $l$, $w$, $h$, $r$ and $s$ represent the length, width, height, rotation angle and classification score of predicted object, respectively. $d$ indicates the combination of all attributes. }
    \label{avg_map}
\end{figure}

\textbf{Average Saliency Map.}  
To further explore the detection mode of detectors and verify the interpretability of FFAM, we average the saliency maps of specific classes, including cars, pedestrians and cyclists. We use SECOND trained on KITTI \cite{kitti} as the detector. To accomplish this, we first scale all boxes and associated points to a uniform size and then align them with respect to their center and rotation angle. Next, we voxelize the resulting point cloud and calculate the average saliency values of individual points within each voxel. The resulting saliency maps for different object attributes are presented in Figure~\ref{avg_map}. As depicted in the first two rows of Figure~\ref{avg_map}, the detector primarily identifies and localizes car objects based on the points located at the four corners of the car. By analyzing features from these points, the detector infers various attributes of a car, such as its center location, length, width, rotation angle, and classification score. This can be attributed to the fact that car objects are often incomplete in outdoor point clouds, and their corners are frequently scanned by LiDAR and used as key features. However, there is a special case, as shown in the first two rows of the penultimate column of Figure~\ref{avg_map}, where the height attribute is predicted primarily based on the points at the top of the car. As illustrated in the third row of Figure~\ref{avg_map}, the detector predicts pedestrian objects mostly based on the points distributed on the head and shoulder regions. Additionally, the detector recognizes cyclist objects mainly based on the points distributed on the head and back of the human body, as shown in the last row of Figure~\ref{avg_map}. Furthermore, we observe that the prediction of cyclist height heavily relies on the points distributed on the head, similar to the prediction of car height. Additional average saliency maps of other detectors are in App.~\ref{vis_maps}.

\subsection{Quantitative Results}
We adopt Deletion, Insertion \cite{rise, drise, odam}, visual explanation accuracy (VEA) \cite{VEA} and Pointing games (PG) to evaluate our FFAM. SECOND trained on KITTI is used as the baseline detector. Following previous work \cite{odam}, we use the well-detected objects in the evaluation dataset as the subjects to be explained. In particular, a predicted object is considered well-detected if the IoU between it and its ground truth is greater than [0,7, 0.5, 0.5] for car, pedestrian and cyclist classes, respectively. See App.~\ref{waymo_quan} for results on Waymo Open.

\begin{figure}
    \begin{minipage}[h]{.5\linewidth}
	\centering
	\includegraphics[scale=0.52]{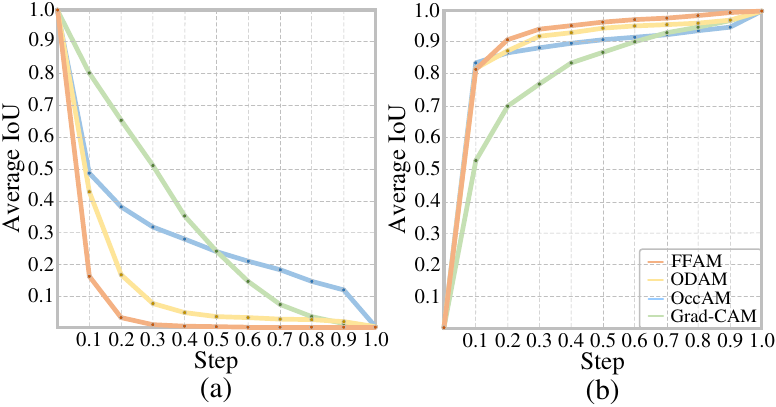} 
	\caption{AUC diagrams for Deletion and Insertion. Average IoU vs. (a) Deletion steps and (b) Insertion steps.}
	\label{quan_plot}
    \end{minipage}
    \hfill
    \begin{minipage}[h]{.45\linewidth}
	\centering
        \resizebox{\linewidth}{!}{
	\begin{tabular}{ccccc}
		\hline
            \multirow{2}{*}{Method} &\multicolumn{4}{c}{VEA $\uparrow$} \\
            \cline{2-5}
            & All & Car & Ped. & Cyc. \\
            \hline
            Grad-CAM & 0.015 & 0.015 & 0.018 & 0.013\\
            ODAM & 0.179 & 0.163 & 0.280 & 0.233 \\
            OccAM & 0.064 & 0.063 & 0.080 & 0.042\\
            FFAM (Ours) & \textbf{0.391} & \textbf{0.363} & \textbf{0.543} & \textbf{0.515}\\
            \hline
	\end{tabular}
        }
        \captionsetup{type=table}
	\caption{Comparison of visual explanation accuracy metric for different categories. `all' denotes the three categories are included. }
        \label{table1}
    \end{minipage}    
\end{figure}

\begin{table}[t]
    \centering
    \begin{tabular}{cccccccccc}
    \hline
    \multirow{2}{*}{Method} &\multicolumn{4}{c}{Deletion $\downarrow$} && \multicolumn{4}{c}{Insertion $\uparrow$} \\
    \cline{2-5} \cline{7-10}
    & All & Car & Ped. & Cyc. && All & Car & Ped. & Cyc.\\
    \hline
    Grad-CAM & 0.335 & 0.373 & 0.137 & 0.129 && 0.797& 0.821 & 0.688 & 0.725\\
    ODAM & 0.134 & 0.138 & 0.122 & 0.098 && 0.885 & 0.902 & 0.785 & 0.828\\
    OccAM & 0.286 & 0.311 & 0.146 & 0.167 && 0.863& 0.880 & 0.761 & 0.790\\
    FFAM (Ours) & \textbf{0.071} & \textbf{0.068} & \textbf{0.098} & \textbf{0.078} && \textbf{0.907} & \textbf{0.923} & \textbf{0.806} & \textbf{0.854}\\
    \hline
    \end{tabular}
    \caption{AUC for Deletion and Insertion curves. The results of different categories are reported. `all' means the combination of the three categories. }
\label{table2}
\end{table}

\textbf{Deletion and Insertion. } 
Deletion and Insertion are widely used to evaluate explanation methods for image-based detection models \cite{drise, odam}. Deletion involves sequentially removing highly salient elements from a scene, measuring the rate model predictions diverge from the original. Insertion progressively adds salient elements to an empty scene, measuring how quickly predictions approach the original. Considering the similarity between pixels in an image and points in a point cloud, we employ Deletion and Insertion to evaluate FFAM. In outdoor point cloud scenes, objects are relatively small compared to global scenes, so we only operate on points within twice the diagonal length of an object's bounding box from its center. We use IoU between a prediction and ground truth as the measure score. Average IoU curves are presented in Figure~\ref{quan_plot}(a-b), and Table~\ref{table2} reports the area under the curve (AUC) for different categories. A lower Deletion AUC indicates a steeper drop in the IoU score, reflecting a more pronounced impact of removed salient points. Conversely, a higher Insertion AUC suggests a larger increase in the IoU score per step, indicating the significance of added salient points. Our methods have the fastest performance drop and largest increase for Deletion and Insertion, showing points highlighted in our saliency maps have a greater effect on detector predictions than the other methods.

\textbf{Visual Explanation Accuracy. } 
VEA calculates the point-level intersection over union (IoU) between the ground truth masks and saliency maps, which are thresholded at various values. The results of VEA for different object categories can be found in Table~\ref{table1}. Notably, our FFAM achieves the highest VEA scores across all categories, indicating the compactness of the visual explanations generated by FFAM. On the other hand, OccAM and Grad-CAM exhibit lower performance on this metric. OccAM tends to mark a significant number of background points, while Grad-CAM is a class-specific visual explanation method, which may explain their comparatively weak performance.

\textbf{Pointing Game. } 
To further assess the localization capability of FFAM, we present the results of the Pointing game (PG). In this evaluation, a hit is recorded if the point with the highest saliency value falls within the ground truth bounding box, while a miss is counted otherwise. The PG metric measures the accuracy of saliency maps by calculating the ratio of hits to the total number of hits and misses. Furthermore, we report the energy-based PG metric (enPG) proposed in \cite{scorecam}, which considers the energy within the ground truth region compared to the global scene. As shown in Table~\ref{table3}, our FFAM surpasses previous methods on all metrics, indicating its superior ability to focus on the explained object. Notably, Grad-CAM performs poorly on both PG and enPG, which aligns with the VEA results presented in Table~\ref{table1}. This suggests that classification-based explanation methods alone are insufficient for generating meaningful explanations for detectors.

\begin{figure}[t]
    \centering
    \includegraphics[width=0.8\textwidth]{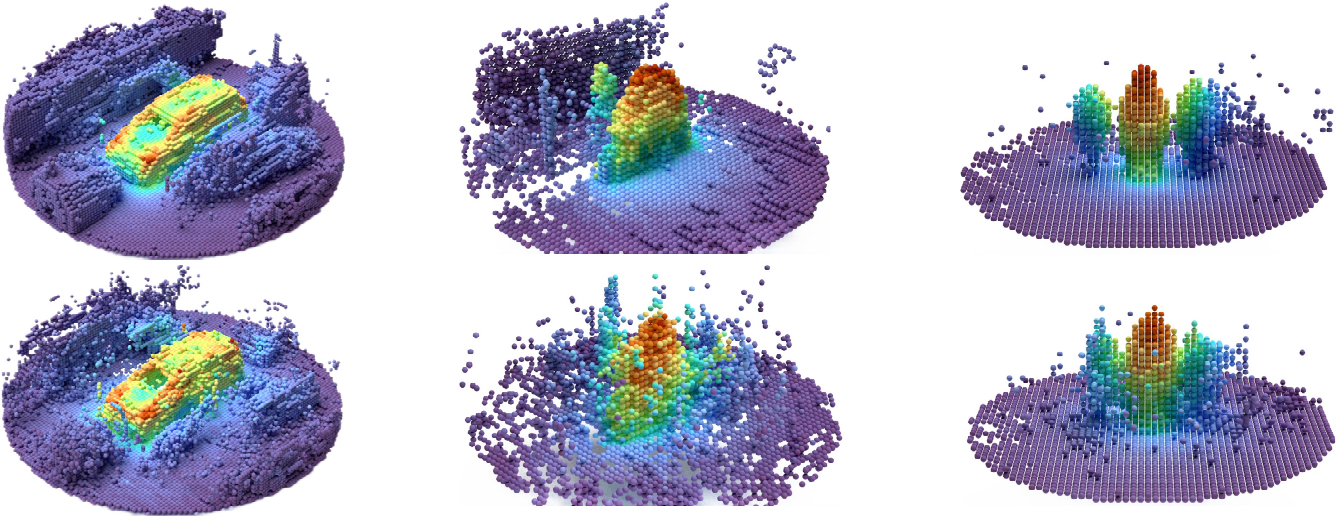}
    \caption{Average saliency maps for true and false positives. The $1^{st}$ and $2^{nd}$ rows represent cases of true and false positives, respectively. }
    \label{failure_average}
\end{figure}

\begin{table}[t]
    \centering
    \begin{tabular}{cccccccccc}
    \hline
    \multirow{2}{*}{Method} &\multicolumn{4}{c}{PG $\uparrow$} && \multicolumn{4}{c}{enPG $\uparrow$} \\
    \cline{2-5} \cline{7-10}
    & All & Car & Ped. & Cyc. && All & Car & Ped. & Cyc.\\
    \hline
    Grad-CAM & 0.093 & 0.080 & 0.166 & 0.163 && 0.021 & 0.022 & 0.014 & 0.011\\
    ODAM & 0.901 & 0.895 & 0.939 & 0.926 && 0.633 & 0.639 & 0.577 & 0.654\\
    OccAM & 0.946 & 0.957 & 0.898 & 0.860 && 0.023 & 0.024 & 0.019 & 0.013\\
    FFAM (Ours) & \textbf{0.991} & \textbf{0.989} & \textbf{0.999} & \textbf{0.998} && \textbf{0.664} & \textbf{0.671} & \textbf{0.591} & \textbf{0.719}\\
    \hline
    \end{tabular}
    \caption{Comparison of Pointing game (PG) and energy-based Pointing game (enPG) metrics.}
\label{table3}
\end{table}

\subsection{Modes of False Positive}
FFAM can be used to identify false positive modes of a detector. A detection is considered a true positive if correctly classified and the Intersection over Union (IoU) between the prediction box and ground truth exceeds a threshold. Otherwise, it is a false positive. The IoU thresholds are 0.7, 0.5, and 0.5 for car, pedestrian, and cyclist objects, aligning with the KITTI official metric \cite{kitti}. To reveal detection modes, we compute average saliency maps separately for true positives and false positives. Results are shown in Figure~\ref{failure_average}. First, we observe the average saliency maps of false positives exhibit similarities to those of true positives. The detector predicts a false positive because it detects a similar pattern to that of a true positive. Second, false positives tend to be surrounded by more noise points, with a point density of approximately one-third of true positives. We believe noises and sparse density may be significant factors contributing to the occurrence of false positives. Lastly, the ratio of car, pedestrian, and cyclist objects in true positives is approximately 36:5:2, while in false positives, it is 13:8:2. This suggests car objects are less prone to false positives compared to pedestrian and cyclist objects.

\section{Conclusion}
In this paper, we propose a visual explanation method, FFAM, that efficiently generates high-quality explanations for 3D detectors. FFAM utilizes non-maximum matrix factorization to obtain a global concept activation map, which is then refined using object-specific gradients. To align the granularity of the input point cloud and intermediate features, we introduce a voxel upsampling strategy. Qualitative and quantitative experiments demonstrate that our FFAM provides more interpretable and compact visual explanations than previous methods. The limitation of FFAM is that it needs to access the feature maps within 3D detectors. In future work, we will explore using visual explanations to enhance the accuracy and efficiency of 3D detectors.

\bibliographystyle{named}
\bibliography{ref}

\clearpage
\appendix
\section{Appendix} 

\begin{table}[htbp]
    \centering
    \begin{tabular}{cccccc}
    \hline
        Layer & Del. $\downarrow$ & Ins. $\uparrow$ & VEA $\uparrow$ & PG $\uparrow$ & enPG $\uparrow$\\
        \hline
        conv1 & 0.167 & 0.911 & 0.084 & 0.885 & 0.562\\
        conv2 & 0.102 & \textbf{0.912} & 0.116 & 0.905 & 0.615\\
        conv3 & \textbf{0.091} & 0.909 & 0.161 & \textbf{0.955} & \textbf{0.654}\\
        conv4 & 0.093 & 0.905 & \textbf{0.208} & 0.946 & 0.644\\
        \hline
    \end{tabular}
    \caption{Results of different layer settings. `conv1', `conv2', `conv3' and `conv4' represent the $1^{st}$, $2^{nd}$, $3^{rd}$ and $4^{th}$ blocks in the 3D backbone. }
    \label{layer_analysis_table}
\end{table}

\begin{table}[htbp]
    \centering
    \begin{tabular}{cccccc}
    \hline
        (Range, $k$) & Del. $\downarrow$ & Ins. $\uparrow$ & VEA $\uparrow$ & PG $\uparrow$ & enPG $\uparrow$\\
        \hline
        (0, 1) & 0.091 & 0.909 & 0.161 & 0.955 & 0.654\\
        (1, 4) & 0.076 & \textbf{0.910} & 0.218 & 0.965 & 0.635\\
        (2, 16) & \textbf{0.069} & 0.909 & \textbf{0.313} & \textbf{0.981} & \textbf{0.644}\\
        (3, 64) & 0.072 & 0.907 & 0.309 & 0.980 & 0.642\\
        \hline
    \end{tabular}
    \caption{Results of different (Range, $k$) settings. `Range' means the Manhattan distance threshold and $k$ denotes the upper bound of neighbor number in the voxel upsampling strategy. }
    \label{range_analysis_table}
\end{table}

\begin{table}[htbp]
    \centering
    \begin{tabular}{cccccc}
    \hline
        $r$ & Del. $\downarrow$ & Ins. $\uparrow$ & VEA $\uparrow$ & PG $\uparrow$ & enPG $\uparrow$\\
        \hline
        8 & \textbf{0.067} & \textbf{0.909} & 0.315 & 0.980 & 0.639\\
        16 & 0.069 & \textbf{0.909} & 0.313 & 0.981 & 0.644\\
        32 & 0.069 & \textbf{0.909} & 0.313 & 0.981 & 0.647\\
        64 & 0.071 & 0.907 & \textbf{0.391} & \textbf{0.991} & \textbf{0.664}\\
        128 & 0.069	& \textbf{0.909} & 0.306 & 0.980 & 0.644\\
        \hline
    \end{tabular}
    \caption{Results of different concept number settings. $r$ denotes the concept number. }
    \label{concept_analysis_table}
\end{table}

\begin{figure}[htbp]
    \centering
    \includegraphics[width=1.0\textwidth]{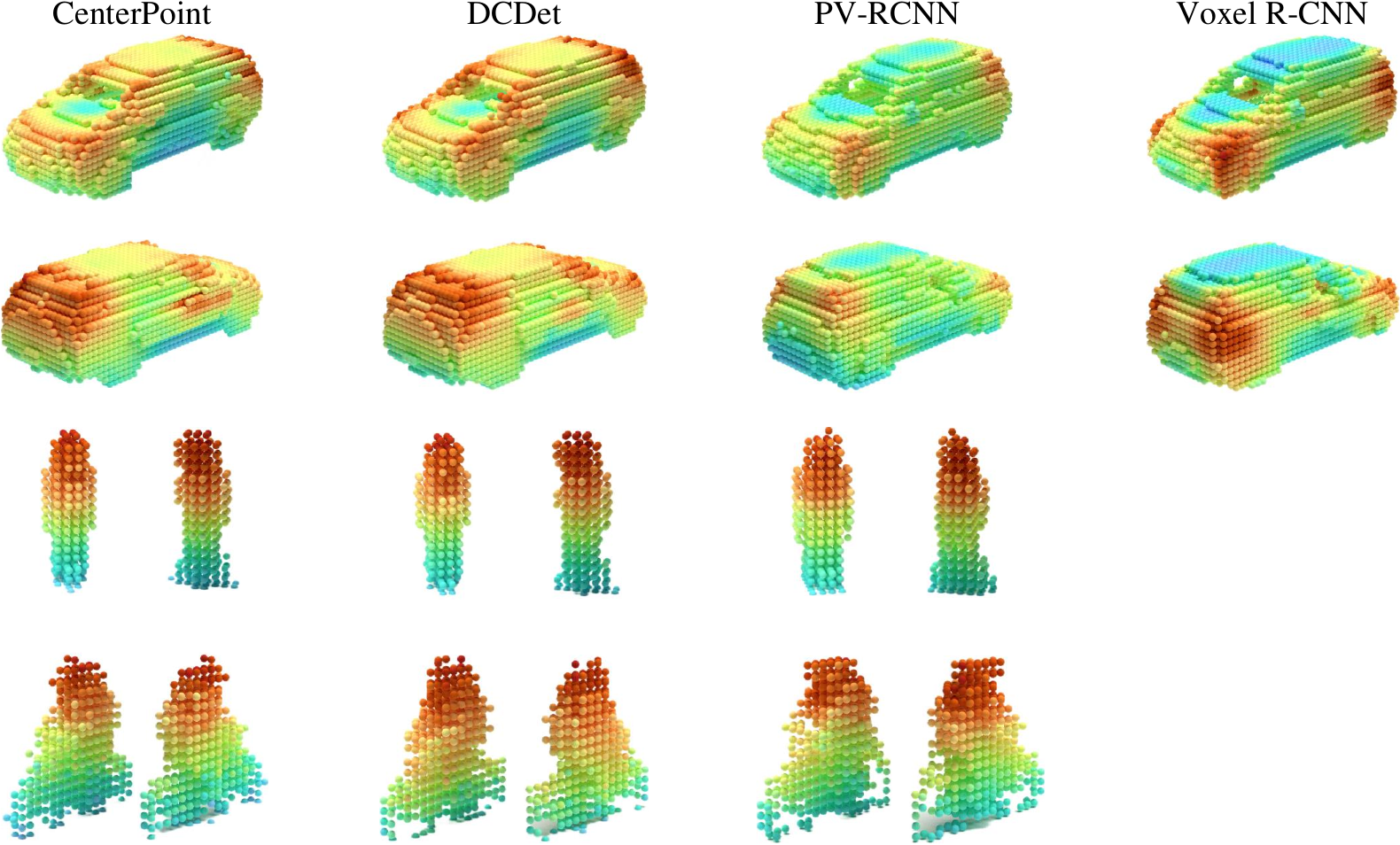}
    \caption{Average saliency maps for different detectors. Voxel R-CNN is only available for the car category (parameter weights are downloaded from OpenPCDet \cite{openpcdet}). }
    \label{other_average}
\end{figure}

\begin{table}[htbp]
    \centering
    \begin{tabular}{cccccccccc}
    \hline
    \multirow{2}{*}{Method} &\multicolumn{4}{c}{Deletion $\downarrow$} && \multicolumn{4}{c}{Insertion $\uparrow$} \\
    \cline{2-5} \cline{7-10}
    & All & Vehicle & Ped. & Cyc. && All & Vehicle & Ped. & Cyc.\\
    \hline
    Grad-CAM & 0.332 & 0.380 & 0.239 & 0.133 && 0.805 & 0.833 & 0.777 & 0.804\\
    ODAM & 0.252 & 0.279 & 0.200 & 0.151 && 0.853 & 0.875 & 0.809 & 0.824\\
    OccAM & 0.562 & 0.594 & 0.505 & 0.357 && 0.871 & 0.893 & 0.826 & 0.832\\
    FFAM (Ours) & \textbf{0.095} & \textbf{0.120} & \textbf{0.085} & \textbf{0.104} && \textbf{0.897} & \textbf{0.922} & \textbf{0.845} & \textbf{0.853}\\
    \hline
    \end{tabular}
    \caption{AUC for Deletion and Insertion curves. The results of different categories are reported. `all' means the combination of the three categories. }
\label{waymo_del_ins}
\end{table}

\begin{table}[htbp]
    \centering
    \begin{tabular}{cccccccccc}
    \hline
    \multirow{2}{*}{Method} &\multicolumn{4}{c}{PG $\uparrow$} && \multicolumn{4}{c}{enPG $\uparrow$} \\
    \cline{2-5} \cline{7-10}
    & All & Car & Ped. & Cyc. && All & Car & Ped. & Cyc.\\
    \hline
    Grad-CAM & 0.028 & 0.016 & 0.045 & 0.247 && 0.006 & 0.008 & 0.002 & 0.004\\
    ODAM & 0.917 & 0.941 & 0.865 & 0.970 && 0.476 & 0.568 & 0.285 & 0.575\\
    OccAM & 0.691 & 0.681 & 0.712 & 0.634 && 0.004 & 0.005 & 0.001 & 0.002\\
    FFAM (Ours) & \textbf{0.975} & \textbf{0.980} & \textbf{0.963} & \textbf{0.980} && \textbf{0.517} & \textbf{0.597} & \textbf{0.349} & \textbf{0.650}\\
    \hline
    \end{tabular}
    \caption{Comparison of Pointing Game and energy-based Pointing Game metrics.}
\label{waymo_pg}
\end{table}

\begin{table}[htbp]
    \centering
    \begin{tabular}{ccccc}
    \hline
        \multirow{2}{*}{Method} &\multicolumn{4}{c}{VEA $\uparrow$} \\
        \cline{2-5}
        & All & Car & Ped. & Cyc. \\
        \hline
        Grad-CAM & 0.009 & 009 & 0.010 & 0.027\\
        ODAM & 0.234 & 0.220 & 0.261 & 0.260 \\
        OccAM & 0.012 & 0.010 & 0.018 & 0.012\\
        FFAM (Ours) & \textbf{0.388} & \textbf{0.358} & \textbf{0.448} & \textbf{0.458}\\
        \hline
    \end{tabular}
    \caption{Comparison of visual explanation accuracy metric for different categories. `all' denotes the three categories are included. }
    \label{waymo_vea}
\end{table}

\begin{table}[t]
    \centering
    \begin{tabular}{cccccccc}
        \hline
        \textit{OG} & \textit{VU} & \textit{FF} & Del. $\downarrow$ & Ins. $\uparrow$ & VEA $\uparrow$ & PG $\uparrow$ & enPG $\uparrow$\\
        \hline
        $\checkmark$ &            & & 0.082 & 0.909 & 0.252 & 0.933 & 0.567 \\
        $\checkmark$ &$\checkmark$& & 0.076 & 0.906 & 0.335 & 0.957 & 0.563 \\
        $\checkmark$ &$\checkmark$&$\checkmark$ & 0.071 & 0.907 & 0.391 & 0.991 & 0.664\\
        \hline
    \end{tabular}
    \caption{Effect of different components of FFAM. \textit{OG}, \textit{VU} and \textit{FF} denote object-specific gradient, voxel upsampling and feature factorization, respectively.}
\label{ablation_study}
\end{table}

\subsection{Hyperparameters Analysis}
\label{hyperp_analysis}
In this section, we determine suitable hyperparameters including the feature map position, concept number $r$ and sampling range for FFAM. Specifically, we select feature maps from the $1^{st}$, $2^{nd}$, $3^{rd}$ and $4^{th}$ blocks of 3D backbone of SECOND, and then generate visual explanations utilizing these feature maps. As shown in Table~\ref{layer_analysis_table}, selecting the feature maps from the $3^{rd}$ block obtains the best results on most metrics. Therefore, in other experiments, we utilize the feature maps from the $3^{rd}$ block. Then, to determine the setting of the sampling range, we set different combinations of the Manhattan distance threshold and neighbor number $k$. As illustrated in Table~\ref{range_analysis_table}, (2, 16) is an appropriate setting for FFAM. Finally, we set $r$ equal to 8, 16, 32, 64 and 128, respectively. The experimental results are shown in Table~\ref{concept_analysis_table}. As we can see, $r=64$ greatly outperforms other settings on VEA, PG and enPG metrics, and performs slightly worse on Deletion and Insertion metrics. Consequently, we select $r=64$ as the default setting.

\subsection{Average Saliency Maps for Other Detectors} 
\label{vis_maps}
To reveal the detection modes of additional detectors, we present average saliency maps from various 3D detectors, namely CenterPoint \cite{centerpoint}, DCDet \cite{dcdet}, PV-RCNN \cite{pv-rcnn}, and Voxel R-CNN \cite{voxelrcnn}. CenterPoint and DCDet are trained and evaluated using Waymo Open \cite{waymo}, while PV-RCNN and Voxel R-CNN employ KITTI \cite{kitti} for training and evaluation. The results are displayed in Figure~\ref{other_average}. Notably, for pedestrian and cyclist categories, different detectors trained on distinct datasets exhibit similar areas of focus, such as the head and shoulder regions for pedestrians and the head and back regions for cyclists. However, in the car/vehicle category, detectors trained on diverse datasets reveal distinct patterns. Detectors trained on Waymo Open tend to concentrate on the front, back, and A-pillars of the vehicle category, whereas detectors trained on KITTI tend to emphasize the four corners of the car category.

\subsection{Quantitative Results on Waymo Open Dataset} 
\label{waymo_quan}
To further assess the visual explanation quality produced by our FFAM, we conducted experiments on Waymo Open \cite{waymo}, utilizing CenterPoint \cite{centerpoint} as the detector for interpretation. We also employ the Deletion, Insertion, PG, enPG and VEA as the evaluation metrics. The results, presented in Table~\ref{waymo_del_ins}-\ref{waymo_vea}, showcase the superior performance of our FFAM across all metrics, mirroring the outcomes observed on KITTI. These findings demonstrate FFAM's remarkable adaptability to diverse detectors trained on different datasets.

\subsection{Ablation Study} 
\label{ablation}
To investigate the impact of each component of FFAM, we conduct an ablation analysis using the SECOND detector on KITTI. Initially, we utilize the object-specific gradient alone to generate saliency maps, which yield relatively satisfactory results, as depicted in the first row of Table~\ref{ablation_study}. Subsequently, we introduce the voxel upsampling strategy into the flow, resulting in significant improvements across most metrics, notably VEA and PG, as indicated in the second row of Table~\ref{ablation_study}. Lastly, we incorporate the complete FFAM components. The third row of Table~\ref{ablation_study} demonstrates that feature factorization greatly enhances the quality of saliency maps, with the PG metric approaching a value close to 1, signifying the precise localization achieved by FFAM. We also observe that voxel upsampling and feature factorization do not yield improvements for the Insertion metric. We believe this is due to the models relying, to some extent, on the neighbor context for detection, whereas voxel upsampling and feature factorization result in a more compact saliency map (i.e., focusing on the object itself).

\end{document}